\pdfoutput=1

\documentclass[11pt]{article}

\usepackage[table]{xcolor}
\usepackage[final]{acl}

\usepackage{times}
\usepackage{latexsym}

\usepackage[T1]{fontenc}

\usepackage[utf8]{inputenc}

\usepackage{microtype}

\usepackage{inconsolata}

\usepackage{graphicx}
\usepackage{amssymb}
\usepackage{pifont}

\usepackage{rotating}
\usepackage{amsmath}
\usepackage{multirow}
\usepackage{paralist}
\usepackage{listings}
\usepackage{makecell} 
\usepackage{booktabs}
\usepackage{multirow}
\usepackage{colortbl}
\usepackage{xcolor}
\usepackage{float}
\definecolor{lightpink}{HTML}{FFF4FA}
\definecolor{darkpink}{HTML}{F881C0}
\definecolor{sagegreen}{HTML}{4DAF4A}
\definecolor{mintgreen}{HTML}{EFF8EE}
\definecolor{lightblue}{HTML}{EDF3F8}
\definecolor{customblue}{HTML}{377EB8}
\definecolor{lightred}{HTML}{FCECEB}
\definecolor{customred}{HTML}{E4191B}

\usepackage[most]{tcolorbox}
\usepackage{wrapfig}
\newtcolorbox{promptboxgreen}[2][]
{
  colframe = sagegreen,
  colback  = mintgreen,
  coltitle = white,  
  title    = {#2},
  #1,
}

\newtcolorbox{promptboxpink}[2][]
{
  colframe = darkpink,
  colback  = lightpink,
  coltitle = white,  
  title    = {#2},
  #1,
}

\newtcolorbox{promptboxblue}[2][]
{
  colframe = customblue,
  colback  = lightblue,
  coltitle = white,  
  title    = {#2},
  #1,
}

\definecolor{customgreen}{HTML}{b2e061}
\title{QG-SMS: Enhancing Test Item Analysis via Student Modeling and Simulation}

\author{
  Bang Nguyen$^{1}$ \quad Tingting Du$^{2}$ \quad Mengxia Yu$^{1}$ \quad Lawrence Angrave$^{3}$ \quad Meng Jiang$^{1}$ \\
  $^{1}$ University of Notre Dame \\
  $^{2}$ University of Wisconsin-Madison \\
  $^{3}$ University of Illinois at Urbana-Champaign \\
\textbf{Correspondence:} \texttt{bnguyen5@nd.edu}
}

\begin{document}
\maketitle
\begin{abstract}

While the Question Generation (QG) task has been increasingly adopted in educational assessments, its evaluation remains limited by approaches that lack a clear connection to the educational values of test items. In this work, we introduce \textit{test item analysis}, a method frequently used by educators to assess test question quality, into QG evaluation. Specifically,  we construct pairs of candidate questions that differ in quality across dimensions such as topic coverage, item difficulty, item discrimination, and distractor efficiency. We then examine whether existing QG evaluation approaches can effectively distinguish these differences. Our findings reveal significant shortcomings in these approaches with respect to accurately assessing test item quality in relation to student performance. To address this gap, we propose a novel QG evaluation framework, QG-SMS, which leverages Large Language Model for Student Modeling and Simulation to perform test item analysis. As demonstrated in our extensive experiments and human evaluation study, the additional perspectives introduced by the simulated student profiles lead to a more effective and robust assessment of test items.
\end{abstract}

\section{Introduction}
\begin{table}[!t]
\centering
\small
\begin{tabular}{@{}p{\linewidth}@{}}
\toprule
    \textbf{Learning Material} \\
    \textbf{Introduction of computer vision}: Computer vision (CV) is the field of computer science that focuses on creating digital systems that can process, analyze, and make sense of visual data [...]. For example, [...] \\
    \textbf{Computer vision history} [...] In 2012, a team from the University of Toronto [...]. The model, called AlexNet, [...],achieved an error rate of 16.4\%, which overperformed all other methods at that time. [...] \\
\midrule
\textbf{Quiz Questions} \\
\textbf{$Q_1$}: Which of the following may utilize computer vision techniques? 1). Use a camera to check potential issues on the surface of products (2). Estimate the freshness of apples from pictures (3). Estimate whether a car is speeding via a camera (4). Determine whether a piece of audio is spoken by a specific person
\\ A) (1)(2)(3);   B) (1)(2)(4); C) (2)(3)(4);   D) (1)(2)(3)(4). \\ \\
\textbf{$Q_2$}: One breakthrough in computer vision happened at the University of Toronto in 2012, which achieved an error rate of [ ] in image classification.
\\ A) 6.4\%; B) 10.4\% C) 12.4\%  D) 16.4\%.\\
 \midrule

\textbf{Evaluation Task: Which question has higher discrimination?} \\
\textbf{\textcolor{customred}{Existing approaches: $Q_1$.}} $Q_1$ is an apply-level question, while $Q_2$ is a recall-level question. \\ 
\textbf{\textcolor{customblue}{Label based on Actual Student Performance: $Q_2$.}} Applications of CV appearing in $Q_1$ can be considered common knowledge while $Q_2$ tests a specific detail which only students who pay close attention to details may be able to answer.
\\
\bottomrule
\end{tabular}
\caption{ Existing LLM-based approaches rely solely on question content for evaluation.
    In this example, ChatEval identifies $Q_1$ as the better test item for distinguishing high- and low-performing students, reasoning that it requires learners to \textit{apply} a concept rather than merely \textit{recall} information  (as in $Q_2$). However, real student performance data shows $Q_1$ has lower discrimination. This highlights the need for evaluation methods that incorporate student modeling. The complete case study is provided in Appendix \ref{app:case-study}.}
\label{tab:motivation}
\end{table}

The Natural Language Processing (NLP) domain has recently seen the growing adoption of the question generation (QG) task in educational assessments to help teachers measure student learning and identify misconceptions \cite{wang2022towards, jia2021eqg, wang-etal-2022-towards, moon-etal-2024-generative, 10.1007/978-3-031-16290-9_20}. These generated questions are often evaluated using reference-based metrics such as ROUGE \cite{lin-2004-rouge}, BLEU \cite{papineni-etal-2002-bleu}, or BERTScore \cite{zhang2019bertscore}, which measure the syntactic and semantic similarity between the generated question and a human-written reference. However, researchers have raised concerns about the validity and reliability of reference-based metrics in accurately reflecting question quality \cite{nguyen-etal-2024-reference}. As a result, reference-free metrics have been proposed to assess aspects of question quality independently of a single reference question \cite{moon-etal-2022-evaluating, nguyen-etal-2024-reference}. Despite these advancements, most reference-free QG metrics primarily focus on the answerability of generated questions, lacking a direct connection to their educational value.


In this work, we introduce \textit{test item analysis}, a well-established method in education for assessing test item quality, into the QG evaluation pipeline. 
In educational testing, test item quality is assessed through both \textit{pre-examination} and \textit{post-examination} analyses. 
Pre-examination analysis evaluates test items (i.e., quiz questions) before administration, focusing on dimensions such as topic alignment, where instructors or subject matter experts ensure that test content aligns with learning objectives \cite{mahjabeen2017difficulty}. 
Post-examination analysis is a powerful tool that evaluates the quality of test questions by analyzing how test takers respond to them. 
It occurs after test administration, providing insights into dimensions such as item difficulty, item discrimination, and distractor efficiency through statistical analyses of test-taker performance \cite{mahjabeen2017difficulty}. Post-examination analysis can help improve future test items' validity and reliability.
However, it cannot evaluate test questions during the test design phase, as it requires test-taker responses that are only available after the test has been administered.

\begin{figure}[t]
    \centering
    \includegraphics[width=\linewidth]{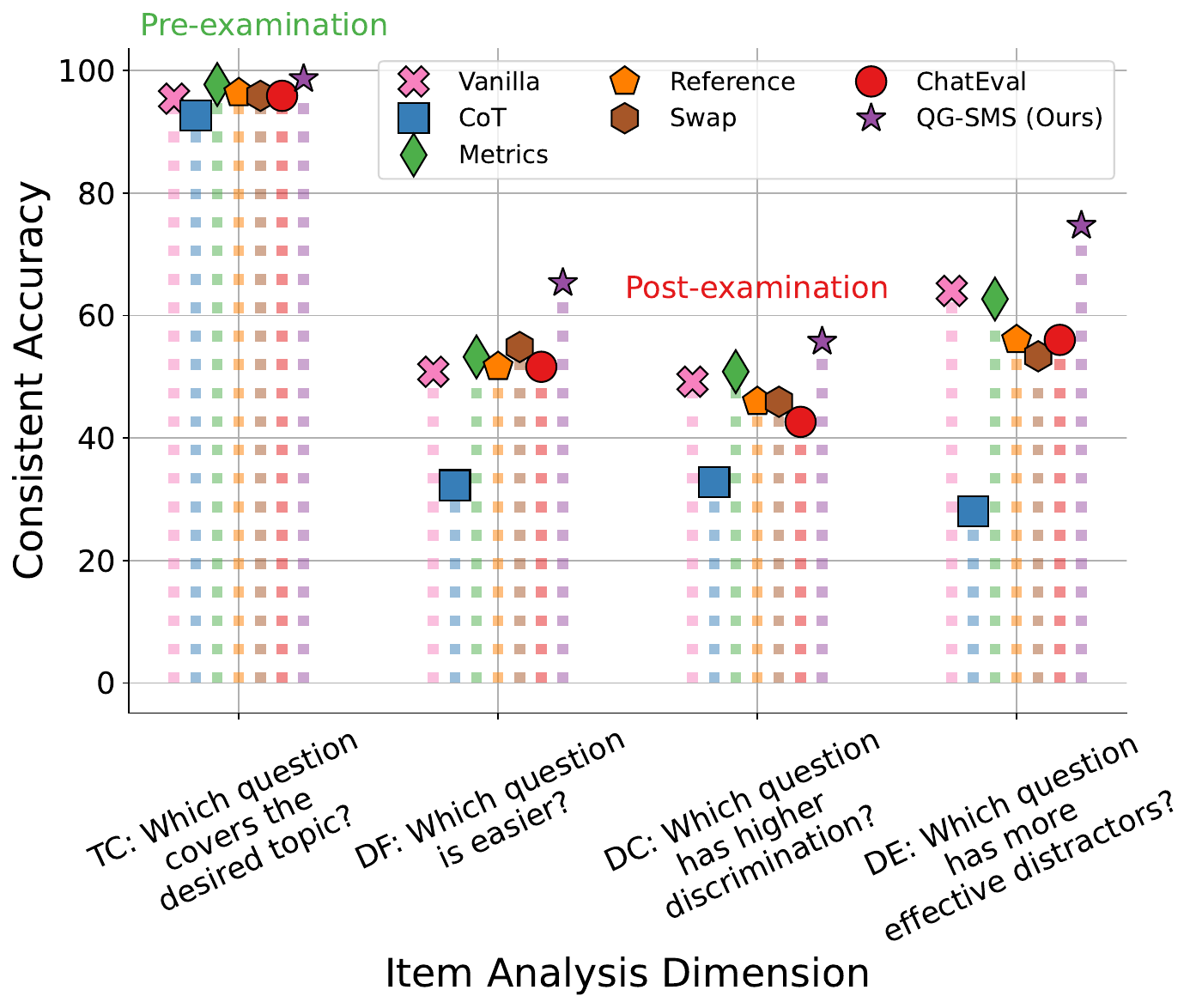}
     \caption{Performance of LLM-based evaluation methods (defined in $\S$\ref{sec:baselines}) in pairwise test item comparisons on the EduAgent dataset.  Existing approaches (in colors except purple/markers except stars) perform well in pre-examination analysis (95.6\%  on average). However, their post-examination performance on question difficulty, discrimination, and distractor efficiency, significantly falls behind, with average consistent accuracies of 49.1\%, 44.5\%, and 53.3\%, respectively. Our proposed approach, QG-SMS, bridges this gap, outperforming all methods across all dimensions.}
    \label{fig:pre_vs_post}
\end{figure}

Recent studies have shown that Large Language Models (LLMs) achieve state-of-the-art alignment with human judgment via pairwise evaluation of generated outputs in natural language generation tasks \cite{chan2023chateval, zeng2024llmbar}. We investigate whether these evaluation approaches can provide a predictive analysis of test items by considering dimensions educators address in both pre-examination and post-examination analyses. Specifically, we consider four dimensions: \textbf{topic coverage} (from pre-examination analysis), and \textbf{item difficulty}, \textbf{item discrimination}, and \textbf{distractor efficiency} (from post-examination analysis). We examine whether existing approaches can effectively distinguish among questions based on these four dimensions--for example, by comparing two questions and identifying which one exhibits higher difficulty. Our findings, illustrated in Fig.~\ref{fig:pre_vs_post}, reveal a significant performance disparity: while existing QG evaluation approaches excel in pre-examination analysis (e.g., topic coverage), they struggle to accurately evaluate dimensions in post-examination analysis, such as item difficulty, discrimination, and distractor efficiency. 


\begin{figure*}[t]
    \centering
    \includegraphics[width=\textwidth]{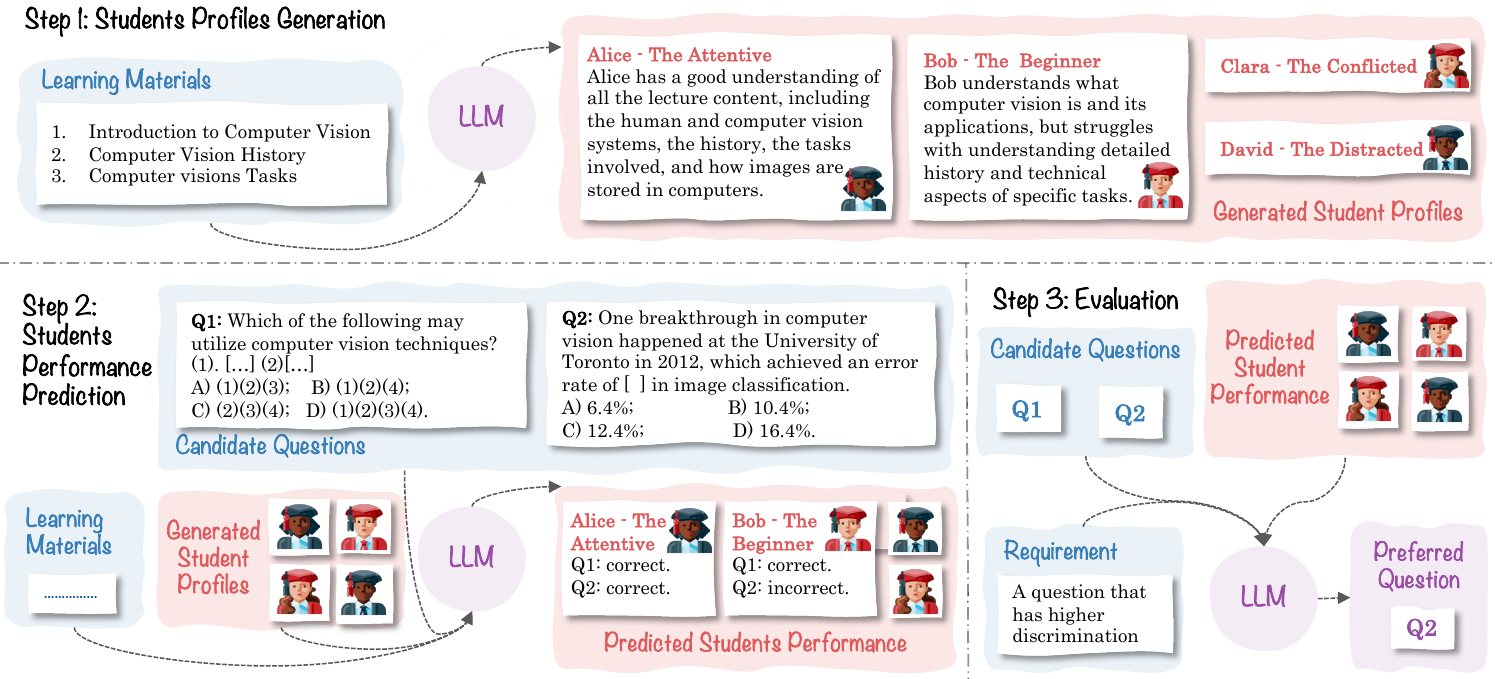}
    \caption{QG-SMS follows three steps: (1) \textbf{Generating student profiles} with diverse understanding of learning materials, (2) \textbf{Predicting responses} of simulated students to candidate questions, and (3) \textbf{Evaluating question quality} based on simulated student performance. In the same example shown in Tbl. \ref{tab:motivation}, QG-SMS arrives at the opposite conclusion from existing evaluation approaches. According to the simulation, applications of computer vision (covered in $Q_1$) are common knowledge among students, including \textit{Alice - The Attentive} and \textit{Bob - The Beginner}, making them equally likely to provide a correct response. Meanwhile, recalling a specific statistic from the lecture (as required by $Q_2$) targets students who pay closer attention like \textit{Alice - The Attentive}. Based on the simulated performance, QG-SMS correctly identifies $Q_2$ as the question with higher discrimination.}
    \label{fig:method-figure}
\end{figure*}

Tbl.~\ref{tab:motivation} illustrates the shortcomings of existing LLM-based evaluation approaches for post-examination analysis. These methods primarily assess question content while neglecting test-taker perspectives, which are crucial for evaluating question quality. To address this gap, we propose \textbf{QG-SMS}, a novel evaluation framework (illustrated in Fig. ~\ref{fig:method-figure}) that utilizes a large language model (LLM) to simulate students with diverse levels of understanding for test item analysis. These simulations serve as reliable indicators of student performance on candidate test items, significantly enhancing the LLM's capacity for evaluating question quality (Fig. ~\ref{fig:pre_vs_post}).
In summary, this paper makes the following contributions:
\begin{itemize}
    \item We systematically introduce \textit{test item analysis} into QG evaluation, revealing a significant performance gap in existing approaches when assessing educational aspects such as question difficulty, discrimination, and distractor efficiency.
    \item To bridge this gap, we propose QG-SMS, a novel QG evaluation framework that leverages diverse \textbf{S}tudent \textbf{M}odeling and \textbf{S}imulation with a single LLM.
    \item We conduct extensive experiments and human evaluation studies to showcase the effectiveness and robustness of QG-SMS.
\end{itemize}

We release all implementation details of QG-SMS to facilitate future works \footnote{\url{https://github.com/bnguyen5/qg-sms}}.

\section{Problem Definition}

\subsection{Statistical Measures of Test Items}
\label{sec:statistics-definiion}
Educators evaluate test items across multiple dimensions to ensure their effectiveness. In this work, we focus on four key dimensions that are well-established in educational research and have been mathematically formalized:
 \textit{topic coverage}, \textit{item difficulty}, \textit{item discrimination}, and \textit{distractor efficiency} \cite{doi:10.3102/0034654309341375, tavakol2011post, mahjabeen2017difficulty}. 
While topic coverage pertains to pre-examination analysis, the remaining dimensions are primarily evaluated post-examination.
 
\textbf{Topic coverage (TC)} evaluates whether the test item covers a given topic. Mathematically, it is a binary variable, where a value of $1$ indicates that the test item covers the desired topic, $0$ otherwise.

\textbf{Item Difficulty (DF)} measures how easy (or difficult) a test item is for a group of students. Let $S = \{s_{1}, ..., s_n\}$ be the set of students who attempted the test item and {$x_s \in \{0,1\}$} indicate whether student $s \in S$ answered correctly. The difficulty index (\textbf{DF}) of the test item is defined as the proportion of students who answered the question correctly:
\[
\textbf{DF} = \frac{\sum_{s \in S} x_s}{|S|}
\]

\textbf{Item Discrimination (DC)} measures the ability of the test item to differentiate between students who have a strong understanding of the learning material and those who do not. Let $X = \{x_{s_1}, x_{s_2}, ..., x_{s_n}\}$ denote the scores of students on the specific test item, and $T = \{t_{s_1}, t_{s_2}, ..., t_{s_n}\}$ where $t_s$ denote the total test score of student $s \in S$. The Discrimination Index \textbf{DC} of the test item is defined as the correlation between the student’s score on the specific item and their overall test score:
\[
\textbf{DC} = \frac{\text{Cov}(X, T)}{\sigma_X \sigma_T}\text{,}
\]
where $\text{Cov}(X, T)$ represents the covariance between $X$ and $T$, while $\sigma_X$, $\sigma_T$ are the standard deviations of $X$ and $T$ respectively.

 For multiple-choice questions, \textbf{distractor efficiency (DE)} assesses how well the distractors (incorrect answer choices) mislead students who hold specific misunderstandings. Let $O$ be the set of distractors of a test item, and $f(s, o) \in \{0,1\}$ denote whether student $s \in S$ selects distractor $o \in O$. Then, the distractor efficiency (\textbf{DE}) of the test item is defined as the number of distractors chosen by at least $5\%$ students in $S$ \cite{mahjabeen2017difficulty}.
\[
\textbf{DE} = |\{o \in O\} | p(o) \geq 0.05|\text{,}
\]
where $p(o) = \frac{\left| \{ s \in S \mid f(s, o) = 1 \} \right|}{|S|}$.

\subsection{Task Definition}
\label{sec:task-def}
Given learning materials $L$ such as lecture content or transcripts, our goal is to obtain a test question that effectively assesses students' knowledge of $L$. Since instructors may have varying requirements for test questions \cite{wang-etal-2022-towards}, let $R_{d}$ denote the desired characteristic or requirement of a test question with respect to a specific dimension $d$ such as question difficulty, discrimination, topic coverage, or distractor efficiency. Given two candidate questions $Q_1$ and $Q_2$ derived from $L$, the task is to determine which question better satisfies the requirement $R_{d}$\footnote{While the current task setup relies on binary comparisons, an extended approach using multiple pairwise comparisons could establish a ranking-based system, where question rankings translate into computed DF/DE/DC scores.}. We provide an example of the task in Tbl. \ref{tab:motivation}.

To ensure that the task is achievable, we require that the statistical measure corresponding to dimension $d$ for $Q_1$ be significantly different from that of $Q_2$. For example, if $d$ represents difficulty, then the absolute difference between the difficulty indices of $Q_1$ and $Q_2$ must exceed a certain threshold $\alpha$: $|\textbf{DF}_{Q_1} - \textbf{DF}_{Q_2}| \geq \alpha$, where $\alpha$ is a predefined threshold ensuring a meaningful distinction between the two questions.

\section{QG-SMS: Student Modeling and Simulation for Test Item Analysis}

During the test design phase, it is imperative to anticipate the diverse ways students may interpret learning materials. For example, in multiple-choice tests, effective distractors help teachers identify students who hold certain misconceptions \cite{doi:10.3102/0034654317726529}. In this sense, to enhance the educational alignment of automated test item evaluation, we propose QG-SMS, which leverages LLM to model and simulate how well test items measure varying levels of student understanding. As illustrated in Fig. \ref{fig:method-figure}, QG-SMS consists of three key steps: 

\textbf{Step 1 - Student Profile Generation}: QG-SMS begins by simulating diverse student perspectives on the same learning materials. Given learning materials $L$, the LLM is tasked to generate a set of students  $S = \{s_{1}, s_2, ..., s_n\}$ such that the distribution of student understanding reflects that in a realistic classroom. 
Note that we only simulate diverse student understanding of the given learning materials, avoiding the use of personal identities that may introduce social bias into the generated profiles \cite{cheng-etal-2023-compost}. Fig.~\ref{fig:method-figure} presents the profiles of two simulated students Alice and Bob.

\textbf{Step 2 - Student Performance Prediction}: Once student profiles are established, QG-SMS simulates their performance on candidate test items. Given learning materials $L$, a pair of candidate questions to be evaluated $\{Q_1, Q_2\}$, and the generated student profiles $S$, the task is to predict whether each student $s \in S$ will correctly or incorrectly answer $Q_1$ and $Q_2$.

\textbf{Step 3 - Evaluation}: Finally, QG-SMS assesses whether a test item fulfills its intended purpose by examining the responses of students with different levels of understanding. For example, an easy question should yield correct answers from a wide range of students, while a challenging question should only be correctly answered by those who have a deeper understanding of the learning materials. In this step, we leverage the LLM's understanding of the question content along with the simulated performance data to make informed judgments on questions. Formally, given the pair of candidate questions $\{Q_1, Q_2\}$, the desired characteristic of the test item $R_d$ and the predicted student performance from step 2, the task is to determine which question better satisfies requirement $R_d$.

Notably, the proposed approach uses the same input $L$, $R_d$, and $\{Q_1, Q_2\}$ as given in $\S$\ref{sec:task-def}. All other information is synthetically simulated by the LLM. We provide the specific prompts used for each step in Appendix. \ref{appendix:prompts}.

\section{Experiments}

\subsection{Dataset Construction}
\label{sec:dataset-construction}

We construct a dataset of question pairs ($Q_1$, $Q_2$) with varying quality levels from two knowledge-tracing datasets: \textit{EduAgent} \citep{xu2024eduagent} and \textit{DBE-KT} \citep{abdelrahman2022dbe} datasets. 
Both datasets contain mappings between learning materials and quiz questions, ensuring that $Q_1$ and $Q_2$ are related to the given learning materials $L$. Each question is also annotated with its relevant topic, allowing us to set up pairs for the topic coverage (\textbf{TC}) setting. In addition, both datasets collect student responses to individual quiz questions, allowing us to compute the statistical measures discussed in $\S$\ref{sec:statistics-definiion}. For \textit{DBE-KT}, we can only compute \textbf{DF} and \textbf{DC} as information on specific distractors chosen by students who answered incorrectly is unavailable.

As discussed in $\S$\ref{sec:task-def}, we adopt the threshold $\alpha$ to ensure a significant quality difference between $Q_1$ and $Q_2$. We set $\alpha$ to $1$ for \textbf{TC}, $2$ for \textbf{DE}, and $0.15$ for \textbf{DF} and \textbf{DC}. For each pair $(Q_1, Q_2)$ that exhibits significant quality difference with respect to dimension $d$, we assign labels based on $d$ and its corresponding requirement $R_d$ as follows:
\begin{itemize}
    \item \textit{Topic coverage}: we define $R_d$ as "the question that covers the target topic". The label corresponds to the question with the higher \textbf{TC} value ($1$ vs $0$).
    \item \textit{Item Difficulty}: we define $R_d$ as "the question that is easier to answer". The label corresponds to the question with the higher \textbf{DF} value.
    \item \textit{Item Discrimination}: we define $R_d$ as "the question that is more effective at distinguishing between high-performing and low-performing students". The label corresponds to the question with the higher \textbf{DC} value.
    \item \textit{Distractor Efficiency}: we define $R_d$ as "the question that has a higher number of effective distractors". The label corresponds to the question with the higher \textbf{DE} value.
\end{itemize}
Notably, $R_d$ can also be defined in the opposite direction to ours without altering the task setup. For example, with difficulty as $d$, $R_d$ can instead be defined as "the question that is more difficult to answer". In this case, the same $(Q_1, Q_2)$ pair would be labeled based on which question has the \textit{lower} \textbf{DF} value.

Ultimately, we obtained $477$ and $255$ question pairs from \textit{EduAgent} and \textit{DBE-KT}, respectively. These pairs serve as a benchmark for evaluating QG-SMS and existing QG evaluation mechanisms across multiple test item dimensions.


\subsection{QG Evaluators}
\label{sec:baselines}
We compare QG-SMS with three \textit{individual-scoring metrics}:


The reference-based  \textbf{BERTScore} \cite{zhang2019bertscore} measures the semantic similarity between the candidate question and a reference. Since we do not have a reference question for each pair, we instead use the learning material $L$ as the reference and measure the similarity between $L$ and each question.

The reference-free \textbf{KDA} \cite{moon-etal-2022-evaluating} evaluates question quality based on the performance of simulated students with and without access to learning material $L$. We use the large version of this model-based metric. 

The LLM-based \textbf{QSalience} \cite{wu-etal-2024-questions} measures the importance of the candidate question for understanding the learning material $L$. We use the best-performing model, \verb|mistral-instruct|, as reported by its authors. 


As these metrics assign separate scores to $Q_1$ and $Q_2$, we must determine how to compare their scores to establish a preference. For each dimension, we select the direction that yields the highest average accuracy for the EduAgent dataset (see Appendix \ref{app:exp-details} for more details) and retain this comparison direction for the DBE-KT dataset, as a reliable metric should exhibit consistent behavior across domains.


\begin{table*}[t]
\centering
\resizebox{\textwidth}{!}{%
\begin{tabular}{lcccc|cccc|cccc|cc}
\toprule
\multirow{4}{*}{\textbf{Method}}
& \multicolumn{4}{c|}{\textbf{Topic Coverage (TC)}} 
& \multicolumn{4}{c|}{\textbf{Difficulty (DF)}} 
& \multicolumn{4}{c|}{\textbf{Discrimination (DC)}} 
& \multicolumn{2}{c}{\textbf{Dist. Eff. (DE)}}  \\
\cmidrule(lr){2-5} \cmidrule(lr){6-9} \cmidrule(lr){10-13} \cmidrule(lr){14-15}
& \multicolumn{2}{c}{\makecell{\textit{EduAgent} \\ 217 pairs}} & \multicolumn{2}{c|}{\makecell{\textit{DBE-KT} \\286 pairs}} 
& \multicolumn{2}{c}{\makecell{\textit{EduAgent} \\ 124 pairs}} & \multicolumn{2}{c|}{\makecell{\textit{DBE-KT} \\162 pairs}} 
& \multicolumn{2}{c}{\makecell{\textit{EduAgent} \\ 61 pairs}} & \multicolumn{2}{c|}{\makecell{\textit{DBE-KT} \\ 93 pairs}} 
& \multicolumn{2}{c}{\makecell{\textit{EduAgent} \\ 75 pairs}} \\
\cmidrule(lr){2-3} \cmidrule(lr){4-5} \cmidrule(lr){6-7} \cmidrule(lr){8-9} \cmidrule(lr){10-11} \cmidrule(lr){12-13} \cmidrule(lr){14-15}
& AA & CA & AA & CA & AA & CA & AA & CA & AA & CA & AA & CA  &  AA & CA \\
\midrule
\textbf{Individual Scoring}  \\
BERTScore       & 79.26 & - & 40.20 & - & 51.61 & - & 61.73 & - & \underline{65.57} & - & 30.11 & - & 65.33 & - \\
KDA$_{large}$&  57.60 & - & 38.46 & - &60.48 & - & 54.32 & - & 60.66 & - & 58.06 & - & \underline{77.33} & -\\
QSalience &  54.84 & - & 48.25 & - & 54.03 & - & 60.49 & - & 52.46 & - & 47.31 & - & \underline{68.00} & -\\

\textbf{Pairwise LLM-based} \\
Vanilla     &  96.54 & 95.39 & 74.30 & 68.89 & 63.71  & 50.80 & 67.28 & 49.38 & 63.11 & 49.18 & 63.98 & 49.46 & 73.33 & \underline{64.00} \\
CoT         &  95.39 & 92.63 & 78.15 & 65.03 &61.69 & 32.26  & 64.20 & 38.89 & 59.84 & 32.79 & 62.90 & 34.41 & 60.00 & 28.00 \\
Metrics     & 97.70 & 97.70 & 80.59 & \textbf{75.17} & 65.32 & 53.22  & 64.20 & 48.77 & \underline{65.57} & \underline{50.82} & 61.29 & 45.16 & 72.00 & 62.67 \\
Reference   & 97.00 & 96.31 & 72.55 & 66.43 & 66.53 & 51.61  & 62.96 & 45.06 & 62.30 & 45.90 & 60.75 & 44.09 & 69.33 & 56.00 \\
Swap        &  95.85 & 95.85 & \textbf{81.64} & 74.48 & 66.53  & \underline{54.84}  & 68.31 & 53.70 & 64.75 & 45.90 & 62.90 & 48.39 & 68.00 & 53.33 \\
ChatEval    &  96.77 & 95.85 & \underline{80.94} & 74.13 &\textbf{68.95} & 51.61 & \textbf{70.99} & \underline{59.88*} & 54.92 & 42.56 & \underline{65.05} & \underline{53.76} & 69.33 & 56.00 \\
\textit{QG-SMS (Ours)} 
            &  \textbf{98.85} & \textbf{98.62} & 79.90 & \underline{74.82} &\underline{68.55} & \textbf{65.32*} & \underline{69.44} & \textbf{64.20*} & \textbf{66.39} & \textbf{55.74} & \textbf{66.66} & \textbf{56.99} & \textbf{79.33} & \textbf{74.67*} \\
\bottomrule
\end{tabular}%
}
\caption{Performance (\textbf{AA}: average accuracy, \textbf{CA}: consistent accuracy) of existing QG evaluation approaches and our proposed QG-SMS approach in test item analysis, grouped by dimension and dataset. The highest and second-highest values for each column are highlighted with \textbf{bold} and \underline{underline} markers, respectively. Asterisks (*) indicate statistical significance at p < 0.1 of LLM-based evaluation approach in improving CA against Vanilla.}
\label{tab:merged_acc}
\end{table*}
We also consider \textit{LLM-based} approaches that perform \textit{pair-wise comparison} of  $Q_1$ and $Q_2$:

\textbf{Vanilla} \cite{zeng2024llmbar}:  We describe the question generation task in natural language, given lecture $L$ and quiz requirement $R_d$, referred to as instruction $I$. Given instruction $I$, the LLM is then asked to choose between $Q_1$ and $Q_2$ based on which question better satisfies $R_d$ (i.e., better aligns with the specified topic, is easier, has higher discrimination ability, or has more effective distractors). The LLM simply outputs its preference without providing an explanation.

\textbf{Chain-of-Thoughts (CoT)} \cite{wei2022chain}: Given instruction $I$, the LLM is prompted to first provide explanations before making its preference between $Q_1$ and $Q_2$.

\textbf{Self-Generated Metrics (Metrics)} \cite{liu-etal-2023-g, saha-etal-2024-branch}: Given instruction $I$, the LLM is first prompted to generate a set of metrics to which a well-constructed test question should adhere. It then selects $Q_1$ or $Q_2$ based on these self-generated metrics.

\textbf{Self-Generated Reference (Reference)} \cite{zheng2023judging}: The LLM is first prompted to generate a reference output (an example of a desirable question) based on instruction $I$. It is then encouraged to utilize this reference to evaluate $Q_1$ and $Q_2$.

\textbf{Swap and Synthesize (Swap)} \cite{10.5555/3692070.3692537}: To address positional bias, the LLM is prompted to express its preference using \textbf{CoT} in both orders $(Q_1, Q_2)$ and $(Q_2, Q_1)$. If the LLM evaluator makes contradictory choices when the question order is swapped, it is prompted to make a final decision by synthesizing the two CoT responses.

\textbf{ChatEval} \cite{chan2023chateval}: This method incorporates multiple personas when using LLM as proxies for human evaluators. Given instruction $I$, we first generate multiple expert personas for the evaluation task using the AutoAgents framework \cite{chen2023autoagents}. The LLM then assumes these personas and engages in a multi-turn discussion to determine its preference between $Q_1$ and $Q_2$.

\subsection{Additional Details}
For all LLM-based evaluation metrics, including ours, we use the same base model, GPT-4o, across all experiments.

As LLMs are known to exhibit strong positional bias \cite{wang-etal-2024-large-language-models-fair}, we run evaluations on each question pair twice, swapping their orders: $(Q_1, Q_2)$ and $(Q_2, Q_1)$. We assess the evaluation performance using two evaluation metrics: \textit{Average Accuracy} and \textit{Consistent Accuracy}. We define \textit{Consistent Accuracy}, applicable to LLM-based methods, as the percentage of cases where the evaluation method makes the correct judgment both when the questions are presented in their original order and when their order is swapped.

Additional experimental details are provided in Appendix \ref{app:exp-details}.

\section{Results}

\begin{figure*}[t]
    \centering
    \includegraphics[width=0.9\textwidth]{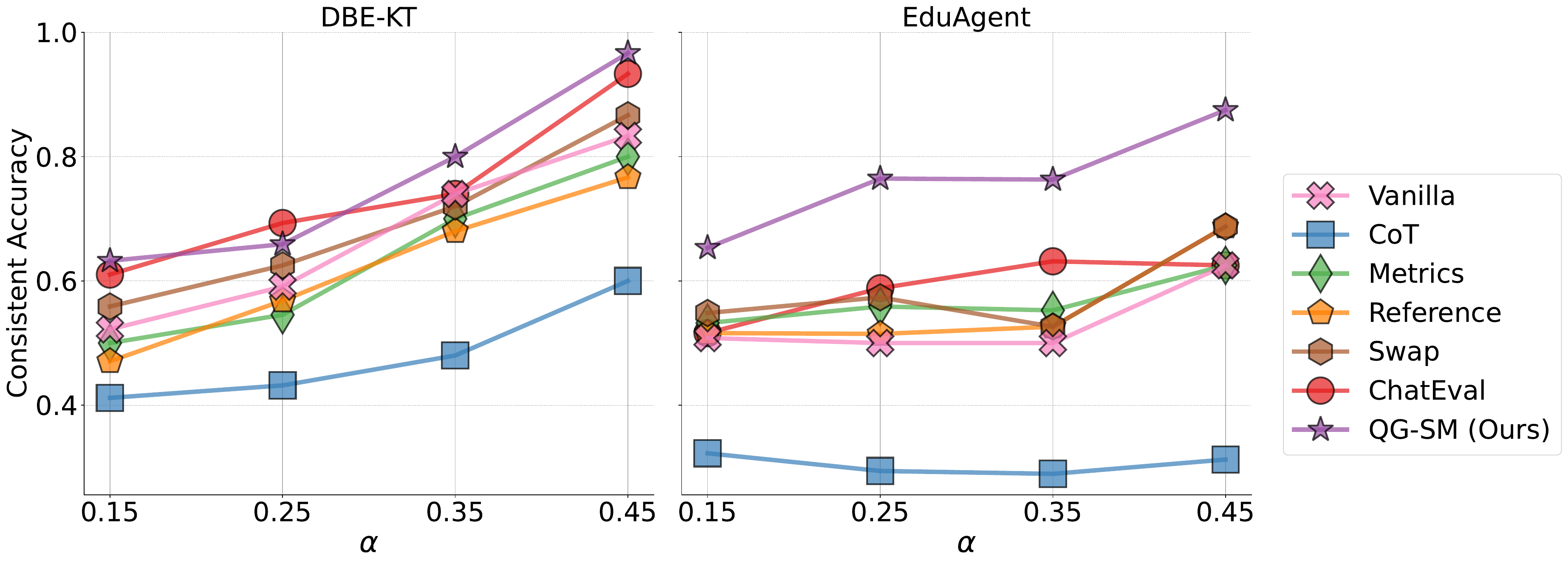}
    \caption{Performance of LLM-based approaches in evaluating for Difficulty (DF) across different $\alpha$ values. QG-SMS consistently shows better evaluation performance compared to other LLM-based approaches.}
    \label{fig:alpha-figure}
\end{figure*}

\subsection{Enhancing Test Item Analysis with QG-SMS}

We provide insights into which dimensions of test item analysis that single-scoring metrics align the most closely, with the expectation that they should achieve accuracy $> 50\%$ on both datasets. As shown in the  Tbl. \ref{tab:merged_acc}, their evaluation behavior is consistent for the DF dimension (i.e., easier questions tend to receive lower BERTScores, higher KDA values, and lower QSalience). KDA is also consistent in its evaluation for DC (higher discrimination questions tend to have lower KDA value), although the evaluation performance is not as comparable to QG-SMS (ours). However, the behavior of BERTScore and QSalience in TC and DC, and KDA in TC, appears dataset-specific and therefore not reliable in reflecting these educational aspects of test items.


While existing LLM-based evaluation approaches perform well in pre-examination analysis of topic coverage (TC), they struggle with post-examination dimensions, as shown in Fig. \ref{fig:pre_vs_post} and Tbl. \ref{tab:merged_acc}. To address this gap, QG-SMS enhances test item analysis performance by incorporating student modeling and simulation. Across both datasets, QG-SMS achieves the highest average accuracy in evaluating DC and DE, and the second-highest average accuracy in evaluating DF. Additionally, QG-SMS significantly outperforms all baselines in consistent accuracy, demonstrating its robustness to input order variations. For instance, QG-SMS's consistent accuracy for DF in the EduAgent dataset is $65.32\%$, maintaining a $10.48\%$ gap over the second-best baseline (Swap). Fig. \ref{fig:method-figure}  provides a case study illustrating how simulation enhances test item analysis, facilitating a more educationally aligned evaluation.

\begin{figure}[t]
    \centering
    \includegraphics[width=\linewidth]{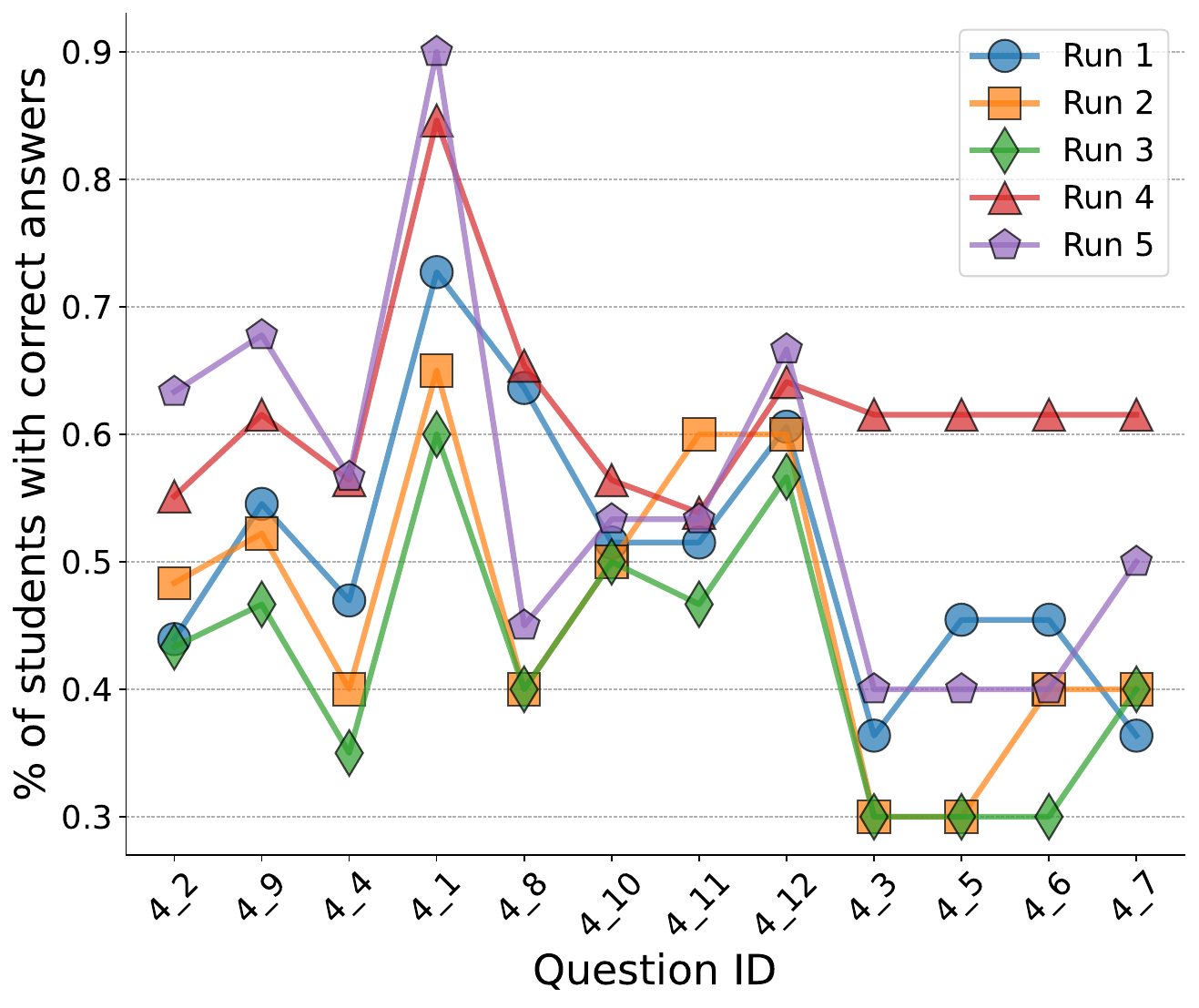}
    \caption{Simulated student performance on the same set of questions across five different runs. The observed consistent distribution of student performance across runs indicates the robustness of the generated student profiles.}
    \label{fig:student_robustness}
\end{figure}

\subsection{Analysis}
\label{sec:analysis}
\textbf{Varying $\alpha$}: We examine the effectiveness of QG-SMS compared to other LLM-based approaches across different values of $\alpha$, i.e., the threshold of quality difference in a pair of questions. Fig. \ref{fig:alpha-figure} indicates that the performance of all LLM-based metrics consistently improves as $\alpha$ increases. This trend is intuitive, as higher $\alpha$ values suggest a larger quality gap in question pairs, making the evaluation task easier. Importantly, QG-SMS remains the top performer regardless of the changes in $\alpha$.

\textbf{Robustness of generated student profiles}: To test the robustness of the generated student profiles, we repeat Step 1 (i.e., student profile generation) and Step 2 (i.e., student performance prediction) multiple times and examine the consistency of the predicted student performance. Fig. \ref{fig:student_robustness} demonstrates that conditioning the student profiles solely on the lecture content already results in consistent distribution of simulated student performance on the same set of questions across different runs. 

\textbf{Necessity of LLM-based evaluation step}: In Step 3, we leverage the same LLM to make preference between candidate question pairs, providing the model with two inputs: the questions' content, and the simulated student performance derived from Step 2. We believe that by providing the LLM with simulated student performances as augmented context, it can more effectively utilize its understanding of the questions' semantics and nuanced language to make informed judgments. To assess the necessity of this step, we perform an ablation study in which we directly compute the Discrimination and Difficulty Index (DC and DF) using the simulated student performance and then compare the question pair based on these statistics. Results indicate that while this direct calculation yields better evaluation performance on Difficulty (with average accuracy improving from $68.55$ to $73.11$), it greatly harms Discrimination performance (with average accuracy decreasing from $66.39$ to $56.83$). This observation demonstrates that Step 3 is effective and necessary for the QG-SMS framework.




\textbf{Extension of QG-SMS to questions without significant quality difference}: We provide this analysis to demonstrate that our pipeline can be applied for test items with similar quality levels. Specifically, we apply QG-SMS to the Distractor Efficiency (DE) dimension within the EduAgent dataset, considering all possible question pairs with $\alpha$ of $0$, $1$, $2$, or $3$. This setting results in a total of 308 pairs. For each unique question, we computed a ranking score as follows: If the question is consistently preferred in both swapped and non-swapped versions of a pair, we add 1 to its score; if it is preferred in only one version, we add 0.5. We then normalize the final score by the total number of pairs the question appears in.

Tbl. \ref{tab:rankq-de} compares QG-SMS-derived scores with actual DE values. An ANOVA test ($p < 0.01$) reveals a significant difference in ranking scores among groups of questions with different DE levels. This supports our claim that QG-SMS can effectively identify groups of questions with similar quality on a specific dimension. We also compare the correlation between these ranking scores and actual DE values across methods. Tbl. \ref{tab:rankq-corr} shows that QG-SMS achieves the highest correlation compared to the two strongest DE baselines (Vanilla and KDA).

\begin{table}[t]
\centering
\resizebox{\linewidth}{!}{%
\begin{tabular}{cccc} 
\toprule
 \textbf{$\text{DE}=0$} & \textbf{$\text{DE}=1$} & \textbf{$\text{DE}=2$}& \textbf{$\text{DE}=3$}\\
\midrule 
0.22 $\pm$ 0.18 & 0.40 $\pm$ 0.22 & 0.48 $\pm$ 0.22 & 0.67 $\pm$ 0.22 \\
\bottomrule
\end{tabular}
}
\caption{QG-SMS-derived ranking scores (mean and standard deviation) of questions with different distractor efficiency (DE) in the EduAgent dataset. Questions with higher DE tend to have higher QG-SMS ranking score.}
\label{tab:rankq-de}
\end{table}

\begin{table}[t]
\centering
\resizebox{0.85\linewidth}{!}{%
\begin{tabular}{lccc}
\toprule
\textbf{Method} & \textbf{Spearman} & \textbf{Kendall} & \textbf{Pearson}\\

\hline
KDA & 0.43 & 0.33 & 0.43  \\
Vanilla & 0.34 & 0.27 & 0.35 \\
QG-SMS & 0.48 & 0.38 & 0.50 \\
\bottomrule
\end{tabular}
}
\caption{Correlation between method for computing ranking score and actual DE value of questions in the EduAgent dataset.}
\label{tab:rankq-corr}
\end{table}

\subsection{Human Evaluation Study}

\begin{table}[t]
\centering
\resizebox{0.8\linewidth}{!}{%
\begin{tabular}{lc|cc}
\toprule
\textbf{Method} & \makecell{\textbf{HumanQs} \\ Stud.Perf Label}
& \multicolumn{2}{c}{\makecell{\textbf{GenQs} \\ Anno Label}}  \\
\cmidrule(lr){2-2} \cmidrule(lr){3-4} 
 & AA & AA & CA  \\

\hline
Vanilla & 70.83 & 70.83 & 58.33  \\
CoT & 67.50 & 65.00 & 38.33  \\
Metrics & 70.83 & 69.17 & 53.33  \\
Reference & 69.17 & 67.50 & 55.00  \\
Swap & 73.33 & 65.00 & 48.33  \\
ChatEval & 69.17 & \textbf{74.17} & \underline{56.67}  \\
QG-SMS & \underline{76.67} & \textbf{74.17} & \textbf{63.33}  \\
Human & \textbf{78.33} & - & -  \\
\bottomrule
\end{tabular}
}
\caption{Results (\textbf{AA}: Average Accuracy, \textbf{CA}: Consistent Accuracy) of QG evaluation approaches on human-written (\textbf{HumanQs}) pairs and generated (\textbf{GenQs}) pairs. The label is determined by actual student performance (\textbf{Stud.Perf}) for the HumanQs pairs, and by Human Annotators (\textbf{Anno}) for the GenQs pairs. The highest and second-highest values for each column are highlighted with \textbf{bold} and \underline{underline} markers, respectively.}
\label{tab:human_eval}
\end{table}
So far, our experiments have involved human-written questions from knowledge-tracing datasets such as \textit{DBE-KT} and \textit{EduAgent}. To further demonstrate the applicability of QG-SMS in the QG process, we conduct a human evaluation study with both human-written and generated questions.

\textbf{Study Description}: We recruit three volunteer annotators, including two graduate and one undergraduate student in Computer Science. Their domain knowledge is highly related to the lecture contents of the \textit{EduAgent} dataset (e.g., AI related knowledge) and they all have some teaching experience. Annotators are tasked to make preferences on $120$ pairs of questions, including $60$ pairs of human-written and $60$ pairs of machine-generated questions. Each pair differs in one of three dimensions - DF, DC, and DE. We use the \textit{EduAgent} dataset. Its lectures target a general audience, supporting the credibility of our annotators in assessing lecture content and quiz questions. We provide more details on the question generation process and instructions given to annotators in Appendix \ref{appendix:human-eval}.

\textbf{Study Results}: In 75 of 120 cases (62.5\%) all three annotators agree on the same preference. For the remaining cases, we adopt the majority preference (chosen by 2 out of 3 annotators) as the representative of human judgment. We report the results of our human evaluation study in Tbl. \ref{tab:human_eval}. 

In human-written question pairs with ground-truth labels based on student performance, our human annotators achieve the highest average accuracy ($78.33\%$) compared to LLM-based evaluators. When broken down by dimension, the average accuracy of human annotators is $90.48\%$, $53.33\%$, and $87.5\%$ for DF, DC, and DE respectively. This observation suggests that performing item analysis on the DC dimension poses significant challenges to our annotators. As they noted during post-examination feedback, it is challenging to identify which question more effectively distinguishes between high-performing and low-performing students when they do not have access to the specific student profiles in the classroom. In terms of evaluating DC, our proposed QG-SMS surpasses human annotators, and on the other two dimensions, DF and DE, QG-SMS achieves the closest accuracy scores to humans. On average, QG-SMS achieves the second-highest accuracy---surpassed only by human annotators. The results show the effectiveness of simulating student understanding and performance. See Tbl.~\ref{tab:human_eval_more} for detailed results.

For the other 60 pairs of generated questions, we use the human annotators' preferences as the labels and evaluate the performance of QG evaluators accordingly. It can be seen from Tbl.~\ref{tab:human_eval} that QG-SMS achieves the highest average accuracy and consistent accuracy in this setting, demonstrating state-of-the-art alignment with human judgment.

\section{Related Work}
\textbf{NLG Evaluation with LLM}: LLM-based evaluators have garnered increasing interest due to their higher correlation with human judgments compared to traditional metrics \citep{zheng2023judging}. As foundation models advance, LLM-based evaluation has evolved from scoring candidate texts based on conditioned probabilities \citep{fu-etal-2024-gptscore} to directly generating scores according to predefined criteria \citep{liu-etal-2023-g}. However, LLMs are sensitive to textual instructions and positional biases. To enhance their reliability, \citet{wang-etal-2024-large-language-models-fair} propose calibration strategies, such as requiring models to generate multiple pieces of evidence and aggregating final scores across different orders of candidates. LLM-based evaluators also benefit from prompting techniques imitating human behaviors such as in-context learning \citep{song-etal-2025-many}, step-by-step reasoning \citep{liu-etal-2023-g}, multi-turn optimization \citep{NEURIPS2023_f64e55d0} and multi-agent debate \citep{chan2023chateval}. Despite these advances, as shown in this work, LLM-based methods still fall short in item analysis, calling for a more effective evaluation strategy like QG-SMS. 

\textbf{Student Modeling and Simulation with (L)LMs}: Recent studies explore the use of (L)LMs to simulate human behaviors in general \cite{10.1145/3586183.3606763}, and classroom learning in particular \cite{xu2023leveraging, zhang2024simulating}. These simulations have been applied in various educational contexts, from training novice teachers \cite{10.1145/3573051.3593393} to promoting student engagement \cite{zhang2024simulating}. 
Prior works have utilized LM-based simulations for evaluating test items. Some limit the simulation to a single group of students \cite{sauberli-clematide-2024-automatic}, while others use multiple (L)LMs with varying capacities to model different students in the classroom \cite{lalor-etal-2019-learning, moon-etal-2022-evaluating, park-etal-2024-large}. Unlike these approaches, our proposed method demonstrates that a single LLM is capable of simulating students at diverse levels, making the pipeline more efficient and scalable. While the approaches proposed by \citet{10.1145/3657604.3662031, lalor-etal-2019-learning, hayakawa2024can, byrd-srivastava-2022-predicting} require manual efforts to control the simulated student profiles through either feature engineering or prompt engineering, our approach eliminates this need, making simulation more flexible. 



\section{Conclusion}
In this work, we proposed QG-SMS, a novel simulation-based QG evaluation framework for test item analysis. We first constructed two datasets of candidate question pairs that differ in quality across multiple dimensions of educational value. Experiments with existing evaluation approaches highlight the challenges of accurately and efficiently assessing test item quality. In response, we introduced the modeling and simulation of diverse student understanding for evaluation. These simulated student profiles offer valuable insights into how well a question functions as a test item for assessing student performance. 

We identify two promising future directions. Prior work has shown that, despite being prompted with educational requirements, LLMs often fail to incorporate them into generated questions, as judged by human evaluators \cite{ALFARABY2024100298}. We have shown that QG-SMS is a reliable indicator of educational aspects like DF, DE, and DC. In this sense, QG-SMS could be integrated into a reward-based optimization pipeline to better align generated test items with educational objectives. Additionally, we observe a growing interest in research question generation \cite{10.1145/3613904.3642698, liu2024personaflow}, which will potentially benefit from a simulation-based evaluation framework like QG-SMS. Existing works still rely on costly and time-consuming evaluation by human researchers. Future work could explore simulating diverse researcher perspectives to enable automated, scalable evaluation of research questions.

\section*{Limitations}
In this work, we evaluate the quality of test items at an individual level. We recognize that constructing assessment typically requires considering multiple dimensions and ensuring diversity within each dimension \cite{osterlind1997constructing}. For example, a well-designed quiz should not only cover different topics from the learning materials rather than repeatedly assessing the same concept, but also include a mix of easy, medium, and hard questions. One potential application of QG-SMS in such scenarios is to rank candidate test items based on a given dimension $d$ by comparing simulated student understanding and performance. Using these rankings, future work could explore methods to assist teachers in assembling assessments that achieve balance across relevant dimensions. Additionally, our significance tests rely on a p-value threshold of $0.1$ (see Appendix \ref{app:exp-details} for more details). Future works could explore whether stronger models could lead to more robust significance results.

\section*{Ethical Considerations}
We avoid introducing bias in the generation and use of student profiles by grounding the simulation in the learning materials alone and instructing the LLM to focus on student understanding, which provides useful insights into test item quality. However, implicit bias may still arise in these generated profiles. For example, despite prompting the LLM to use names that describe student understanding, we observed a predominance of European names (\textit{Alice}, \textit{Bob}, etc.). It is important to emphasize that these simulated profiles are not intended to represent specific students in a real classroom. Rather, they serve collectively to estimate the diversity of student understanding of the learning materials.

\section*{Acknowledgments}
This work was supported by NSF IIS-2119531, IIS-2137396, IIS-2142827, IIS-2234058, CCF-1901059, and ONR N00014-22-1-2507.

\bibliography{custom, anthology}

\begin{thebibliography}{45}
\providecommand{\natexlab}[1]{#1}

\bibitem[{Abdelrahman et~al.(2022)Abdelrahman, Abdelfattah, Wang, and Lin}]{abdelrahman2022dbe}
Ghodai Abdelrahman, Sherif Abdelfattah, Qing Wang, and Yu~Lin. 2022.
\newblock Dbe-kt22: A knowledge tracing dataset based on online student evaluation.
\newblock \emph{arXiv preprint arXiv:2208.12651}.

\bibitem[{{Al Faraby} et~al.(2024){Al Faraby}, Romadhony, and Adiwijaya}]{ALFARABY2024100298}
Said {Al Faraby}, Ade Romadhony, and Adiwijaya. 2024.
\newblock \href {https://doi.org/10.1016/j.caeai.2024.100298} {Analysis of llms for educational question classification and generation}.
\newblock \emph{Computers and Education: Artificial Intelligence}, 7:100298.

\bibitem[{Bai et~al.(2023)Bai, Ying, Cao, Lv, He, Wang, Yu, Zeng, Xiao, Lyu, Zhang, Li, and Hou}]{NEURIPS2023_f64e55d0}
Yushi Bai, Jiahao Ying, Yixin Cao, Xin Lv, Yuze He, Xiaozhi Wang, Jifan Yu, Kaisheng Zeng, Yijia Xiao, Haozhe Lyu, Jiayin Zhang, Juanzi Li, and Lei Hou. 2023.
\newblock \href {https://proceedings.neurips.cc/paper_files/paper/2023/file/f64e55d03e2fe61aa4114e49cb654acb-Paper-Datasets_and_Benchmarks.pdf} {Benchmarking foundation models with language-model-as-an-examiner}.
\newblock In \emph{Advances in Neural Information Processing Systems}, volume~36, pages 78142--78167. Curran Associates, Inc.

\bibitem[{Byrd and Srivastava(2022)}]{byrd-srivastava-2022-predicting}
Matthew Byrd and Shashank Srivastava. 2022.
\newblock \href {https://doi.org/10.18653/v1/2022.acl-short.15} {Predicting difficulty and discrimination of natural language questions}.
\newblock In \emph{Proceedings of the 60th Annual Meeting of the Association for Computational Linguistics (Volume 2: Short Papers)}, pages 119--130, Dublin, Ireland. Association for Computational Linguistics.

\bibitem[{Chan et~al.(2023)Chan, Chen, Su, Yu, Xue, Zhang, Fu, and Liu}]{chan2023chateval}
Chi-Min Chan, Weize Chen, Yusheng Su, Jianxuan Yu, Wei Xue, Shanghang Zhang, Jie Fu, and Zhiyuan Liu. 2023.
\newblock Chateval: Towards better llm-based evaluators through multi-agent debate.
\newblock \emph{arXiv preprint arXiv:2308.07201}.

\bibitem[{Chen et~al.(2023)Chen, Dong, Shu, Zhang, Sesay, Karlsson, Fu, and Shi}]{chen2023autoagents}
Guangyao Chen, Siwei Dong, Yu~Shu, Ge~Zhang, Jaward Sesay, B{\"o}rje~F Karlsson, Jie Fu, and Yemin Shi. 2023.
\newblock Autoagents: A framework for automatic agent generation.
\newblock \emph{arXiv preprint arXiv:2309.17288}.

\bibitem[{Cheng et~al.(2023)Cheng, Piccardi, and Yang}]{cheng-etal-2023-compost}
Myra Cheng, Tiziano Piccardi, and Diyi Yang. 2023.
\newblock \href {https://doi.org/10.18653/v1/2023.emnlp-main.669} {{C}o{MP}os{T}: Characterizing and evaluating caricature in {LLM} simulations}.
\newblock In \emph{Proceedings of the 2023 Conference on Empirical Methods in Natural Language Processing}, pages 10853--10875, Singapore. Association for Computational Linguistics.

\bibitem[{Du et~al.(2024)Du, Li, Torralba, Tenenbaum, and Mordatch}]{10.5555/3692070.3692537}
Yilun Du, Shuang Li, Antonio Torralba, Joshua~B. Tenenbaum, and Igor Mordatch. 2024.
\newblock Improving factuality and reasoning in language models through multiagent debate.
\newblock In \emph{Proceedings of the 41st International Conference on Machine Learning}, ICML'24. JMLR.org.

\bibitem[{Fu et~al.(2024)Fu, Ng, Jiang, and Liu}]{fu-etal-2024-gptscore}
Jinlan Fu, See-Kiong Ng, Zhengbao Jiang, and Pengfei Liu. 2024.
\newblock \href {https://doi.org/10.18653/v1/2024.naacl-long.365} {{GPTS}core: Evaluate as you desire}.
\newblock In \emph{Proceedings of the 2024 Conference of the North American Chapter of the Association for Computational Linguistics: Human Language Technologies (Volume 1: Long Papers)}, pages 6556--6576, Mexico City, Mexico. Association for Computational Linguistics.

\bibitem[{Gierl et~al.(2017)Gierl, Bulut, Guo, and Zhang}]{doi:10.3102/0034654317726529}
Mark~J. Gierl, Okan Bulut, Qi~Guo, and Xinxin Zhang. 2017.
\newblock \href {https://doi.org/10.3102/0034654317726529} {Developing, analyzing, and using distractors for multiple-choice tests in education: A comprehensive review}.
\newblock \emph{Review of Educational Research}, 87(6):1082--1116.

\bibitem[{Hayakawa and Saggion(2024)}]{hayakawa2024can}
Akio Hayakawa and Horacio Saggion. 2024.
\newblock Can llms solve reading comprehension tests as second language learners?
\newblock In \emph{Fourth Workshop on Knowledge-infused Learning}.

\bibitem[{Jia et~al.(2021)Jia, Zhou, Sun, and Wu}]{jia2021eqg}
Xin Jia, Wenjie Zhou, Xu~Sun, and Yunfang Wu. 2021.
\newblock Eqg-race: Examination-type question generation.
\newblock In \emph{Proceedings of the AAAI conference on artificial intelligence}, volume~35, pages 13143--13151.

\bibitem[{Lalor et~al.(2019)Lalor, Wu, and Yu}]{lalor-etal-2019-learning}
John~P. Lalor, Hao Wu, and Hong Yu. 2019.
\newblock \href {https://doi.org/10.18653/v1/D19-1434} {Learning latent parameters without human response patterns: Item response theory with artificial crowds}.
\newblock In \emph{Proceedings of the 2019 Conference on Empirical Methods in Natural Language Processing and the 9th International Joint Conference on Natural Language Processing (EMNLP-IJCNLP)}, pages 4249--4259, Hong Kong, China. Association for Computational Linguistics.

\bibitem[{Lin(2004)}]{lin-2004-rouge}
Chin-Yew Lin. 2004.
\newblock \href {https://aclanthology.org/W04-1013/} {{ROUGE}: A package for automatic evaluation of summaries}.
\newblock In \emph{Text Summarization Branches Out}, pages 74--81, Barcelona, Spain. Association for Computational Linguistics.

\bibitem[{Liu et~al.(2023)Liu, Iter, Xu, Wang, Xu, and Zhu}]{liu-etal-2023-g}
Yang Liu, Dan Iter, Yichong Xu, Shuohang Wang, Ruochen Xu, and Chenguang Zhu. 2023.
\newblock \href {https://doi.org/10.18653/v1/2023.emnlp-main.153} {{G}-eval: {NLG} evaluation using gpt-4 with better human alignment}.
\newblock In \emph{Proceedings of the 2023 Conference on Empirical Methods in Natural Language Processing}, pages 2511--2522, Singapore. Association for Computational Linguistics.

\bibitem[{Liu et~al.(2024{\natexlab{a}})Liu, Chen, Cheng, Yu, Ran, Mo, Tang, and Huang}]{10.1145/3613904.3642698}
Yiren Liu, Si~Chen, Haocong Cheng, Mengxia Yu, Xiao Ran, Andrew Mo, Yiliu Tang, and Yun Huang. 2024{\natexlab{a}}.
\newblock \href {https://doi.org/10.1145/3613904.3642698} {How ai processing delays foster creativity: Exploring research question co-creation with an llm-based agent}.
\newblock In \emph{Proceedings of the 2024 CHI Conference on Human Factors in Computing Systems}, CHI '24, New York, NY, USA. Association for Computing Machinery.

\bibitem[{Liu et~al.(2024{\natexlab{b}})Liu, Sharma, Oswal, Xia, and Huang}]{liu2024personaflow}
Yiren Liu, Pranav Sharma, Mehul~Jitendra Oswal, Haijun Xia, and Yun Huang. 2024{\natexlab{b}}.
\newblock Personaflow: Boosting research ideation with llm-simulated expert personas.
\newblock \emph{arXiv preprint arXiv:2409.12538}.

\bibitem[{Lu and Wang(2024)}]{10.1145/3657604.3662031}
Xinyi Lu and Xu~Wang. 2024.
\newblock \href {https://doi.org/10.1145/3657604.3662031} {Generative students: Using llm-simulated student profiles to support question item evaluation}.
\newblock In \emph{Proceedings of the Eleventh ACM Conference on Learning @ Scale}, L@S '24, page 16–27, New York, NY, USA. Association for Computing Machinery.

\bibitem[{Mahjabeen et~al.(2017)Mahjabeen, Alam, Hassan, Zafar, Butt, Konain, and Rizvi}]{mahjabeen2017difficulty}
Wajiha Mahjabeen, Saeed Alam, Usman Hassan, Tahira Zafar, Rubab Butt, Sadaf Konain, and Myedah Rizvi. 2017.
\newblock Difficulty index, discrimination index and distractor efficiency in multiple choice questions.
\newblock \emph{Annals of PIMS-Shaheed Zulfiqar Ali Bhutto Medical University}, 13(4):310--315.

\bibitem[{Markel et~al.(2023)Markel, Opferman, Landay, and Piech}]{10.1145/3573051.3593393}
Julia~M. Markel, Steven~G. Opferman, James~A. Landay, and Chris Piech. 2023.
\newblock \href {https://doi.org/10.1145/3573051.3593393} {Gpteach: Interactive ta training with gpt-based students}.
\newblock In \emph{Proceedings of the Tenth ACM Conference on Learning @ Scale}, L@S '23, page 226–236, New York, NY, USA. Association for Computing Machinery.

\bibitem[{Martone and Sireci(2009)}]{doi:10.3102/0034654309341375}
Andrea Martone and Stephen~G. Sireci. 2009.
\newblock \href {https://doi.org/10.3102/0034654309341375} {Evaluating alignment between curriculum, assessment, and instruction}.
\newblock \emph{Review of Educational Research}, 79(4):1332--1361.

\bibitem[{Moon et~al.(2022)Moon, Yang, Yu, Lee, Jeong, Park, Shin, Kim, and Choi}]{moon-etal-2022-evaluating}
Hyeongdon Moon, Yoonseok Yang, Hangyeol Yu, Seunghyun Lee, Myeongho Jeong, Juneyoung Park, Jamin Shin, Minsam Kim, and Seungtaek Choi. 2022.
\newblock \href {https://doi.org/10.18653/v1/2022.emnlp-main.718} {Evaluating the knowledge dependency of questions}.
\newblock In \emph{Proceedings of the 2022 Conference on Empirical Methods in Natural Language Processing}, pages 10512--10526, Abu Dhabi, United Arab Emirates. Association for Computational Linguistics.

\bibitem[{Moon et~al.(2024)Moon, Lee, Eo, Park, Seo, and Lim}]{moon-etal-2024-generative}
Hyeonseok Moon, Jaewook Lee, Sugyeong Eo, Chanjun Park, Jaehyung Seo, and Heuiseok Lim. 2024.
\newblock \href {https://aclanthology.org/2024.findings-eacl.145/} {Generative interpretation: Toward human-like evaluation for educational question-answer pair generation}.
\newblock In \emph{Findings of the Association for Computational Linguistics: EACL 2024}, pages 2185--2196, St. Julian{'}s, Malta. Association for Computational Linguistics.

\bibitem[{Nguyen et~al.(2024)Nguyen, Yu, Huang, and Jiang}]{nguyen-etal-2024-reference}
Bang Nguyen, Mengxia Yu, Yun Huang, and Meng Jiang. 2024.
\newblock \href {https://doi.org/10.18653/v1/2024.findings-emnlp.798} {Reference-based metrics disprove themselves in question generation}.
\newblock In \emph{Findings of the Association for Computational Linguistics: EMNLP 2024}, pages 13651--13666, Miami, Florida, USA. Association for Computational Linguistics.

\bibitem[{Nguyen et~al.(2022)Nguyen, Bhat, Moore, Bier, and Stamper}]{10.1007/978-3-031-16290-9_20}
Huy~A. Nguyen, Shravya Bhat, Steven Moore, Norman Bier, and John Stamper. 2022.
\newblock \href {https://doi.org/10.1007/978-3-031-16290-9_20} {Towards generalized methods for automatic question generation in educational domains}.
\newblock In \emph{Educating for a New Future: Making Sense of Technology-Enhanced Learning Adoption: 17th European Conference on Technology Enhanced Learning, EC-TEL 2022, Toulouse, France, September 12–16, 2022, Proceedings}, page 272–284, Berlin, Heidelberg. Springer-Verlag.

\bibitem[{Osterlind(1997)}]{osterlind1997constructing}
S.J. Osterlind. 1997.
\newblock \href {https://books.google.com/books?id=Ia3SGDfbaV0C} {\emph{Constructing Test Items: Multiple-Choice, Constructed-Response, Performance and Other Formats}}.
\newblock Evaluation in Education and Human Services. Springer Netherlands.

\bibitem[{Pang et~al.(2024)Pang, Tang, Ye, Xiong, Zhang, Wang, and Chen}]{10.5555/3692070.3693667}
Xianghe Pang, Shuo Tang, Rui Ye, Yuxin Xiong, Bolun Zhang, Yanfeng Wang, and Siheng Chen. 2024.
\newblock Self-alignment of large language models via monopolylogue-based social scene simulation.
\newblock In \emph{Proceedings of the 41st International Conference on Machine Learning}, ICML'24. JMLR.org.

\bibitem[{Papineni et~al.(2002)Papineni, Roukos, Ward, and Zhu}]{papineni-etal-2002-bleu}
Kishore Papineni, Salim Roukos, Todd Ward, and Wei-Jing Zhu. 2002.
\newblock \href {https://doi.org/10.3115/1073083.1073135} {{B}leu: a method for automatic evaluation of machine translation}.
\newblock In \emph{Proceedings of the 40th Annual Meeting of the Association for Computational Linguistics}, pages 311--318, Philadelphia, Pennsylvania, USA. Association for Computational Linguistics.

\bibitem[{Park et~al.(2024)Park, Park, Won, and Kim}]{park-etal-2024-large}
Jae-Woo Park, Seong-Jin Park, Hyun-Sik Won, and Kang-Min Kim. 2024.
\newblock \href {https://doi.org/10.18653/v1/2024.findings-emnlp.477} {Large language models are students at various levels: Zero-shot question difficulty estimation}.
\newblock In \emph{Findings of the Association for Computational Linguistics: EMNLP 2024}, pages 8157--8177, Miami, Florida, USA. Association for Computational Linguistics.

\bibitem[{Park et~al.(2023)Park, O'Brien, Cai, Morris, Liang, and Bernstein}]{10.1145/3586183.3606763}
Joon~Sung Park, Joseph O'Brien, Carrie~Jun Cai, Meredith~Ringel Morris, Percy Liang, and Michael~S. Bernstein. 2023.
\newblock \href {https://doi.org/10.1145/3586183.3606763} {Generative agents: Interactive simulacra of human behavior}.
\newblock In \emph{Proceedings of the 36th Annual ACM Symposium on User Interface Software and Technology}, UIST '23, New York, NY, USA. Association for Computing Machinery.

\bibitem[{Saha et~al.(2024)Saha, Levy, Celikyilmaz, Bansal, Weston, and Li}]{saha-etal-2024-branch}
Swarnadeep Saha, Omer Levy, Asli Celikyilmaz, Mohit Bansal, Jason Weston, and Xian Li. 2024.
\newblock \href {https://doi.org/10.18653/v1/2024.naacl-long.462} {Branch-solve-merge improves large language model evaluation and generation}.
\newblock In \emph{Proceedings of the 2024 Conference of the North American Chapter of the Association for Computational Linguistics: Human Language Technologies (Volume 1: Long Papers)}, pages 8352--8370, Mexico City, Mexico. Association for Computational Linguistics.

\bibitem[{S{\"a}uberli and Clematide(2024)}]{sauberli-clematide-2024-automatic}
Andreas S{\"a}uberli and Simon Clematide. 2024.
\newblock \href {https://aclanthology.org/2024.readi-1.3/} {Automatic generation and evaluation of reading comprehension test items with large language models}.
\newblock In \emph{Proceedings of the 3rd Workshop on Tools and Resources for People with REAding DIfficulties (READI) @ LREC-COLING 2024}, pages 22--37, Torino, Italia. ELRA and ICCL.

\bibitem[{Song et~al.(2025)Song, Zheng, and Luo}]{song-etal-2025-many}
Mingyang Song, Mao Zheng, and Xuan Luo. 2025.
\newblock \href {https://aclanthology.org/2025.coling-main.548/} {Can many-shot in-context learning help {LLM}s as evaluators? a preliminary empirical study}.
\newblock In \emph{Proceedings of the 31st International Conference on Computational Linguistics}, pages 8232--8241, Abu Dhabi, UAE. Association for Computational Linguistics.

\bibitem[{Tavakol and Dennick(2011)}]{tavakol2011post}
Mohsen Tavakol and Reg Dennick. 2011.
\newblock Post-examination analysis of objective tests.
\newblock \emph{Medical teacher}, 33(6):447--458.

\bibitem[{Wang et~al.(2024)Wang, Li, Chen, Cai, Zhu, Lin, Cao, Kong, Liu, Liu, and Sui}]{wang-etal-2024-large-language-models-fair}
Peiyi Wang, Lei Li, Liang Chen, Zefan Cai, Dawei Zhu, Binghuai Lin, Yunbo Cao, Lingpeng Kong, Qi~Liu, Tianyu Liu, and Zhifang Sui. 2024.
\newblock \href {https://doi.org/10.18653/v1/2024.acl-long.511} {Large language models are not fair evaluators}.
\newblock In \emph{Proceedings of the 62nd Annual Meeting of the Association for Computational Linguistics (Volume 1: Long Papers)}, pages 9440--9450, Bangkok, Thailand. Association for Computational Linguistics.

\bibitem[{Wang et~al.(2022{\natexlab{a}})Wang, Fan, Houghton, and Wang}]{wang-etal-2022-towards}
Xu~Wang, Simin Fan, Jessica Houghton, and Lu~Wang. 2022{\natexlab{a}}.
\newblock \href {https://doi.org/10.18653/v1/2022.naacl-main.22} {Towards process-oriented, modular, and versatile question generation that meets educational needs}.
\newblock In \emph{Proceedings of the 2022 Conference of the North American Chapter of the Association for Computational Linguistics: Human Language Technologies}, pages 291--302, Seattle, United States. Association for Computational Linguistics.

\bibitem[{Wang et~al.(2022{\natexlab{b}})Wang, Valdez, Basu~Mallick, and Baraniuk}]{wang2022towards}
Zichao Wang, Jakob Valdez, Debshila Basu~Mallick, and Richard~G Baraniuk. 2022{\natexlab{b}}.
\newblock Towards human-like educational question generation with large language models.
\newblock In \emph{International conference on artificial intelligence in education}, pages 153--166. Springer.

\bibitem[{Wei et~al.(2022)Wei, Wang, Schuurmans, Bosma, Xia, Chi, Le, Zhou et~al.}]{wei2022chain}
Jason Wei, Xuezhi Wang, Dale Schuurmans, Maarten Bosma, Fei Xia, Ed~Chi, Quoc~V Le, Denny Zhou, et~al. 2022.
\newblock Chain-of-thought prompting elicits reasoning in large language models.
\newblock \emph{Advances in neural information processing systems}, 35:24824--24837.

\bibitem[{Wu et~al.(2024)Wu, Mangla, Dimakis, Durrett, and Li}]{wu-etal-2024-questions}
Yating Wu, Ritika~Rajesh Mangla, Alex Dimakis, Greg Durrett, and Junyi~Jessy Li. 2024.
\newblock \href {https://doi.org/10.18653/v1/2024.emnlp-main.1114} {Which questions should {I} answer? salience prediction of inquisitive questions}.
\newblock In \emph{Proceedings of the 2024 Conference on Empirical Methods in Natural Language Processing}, pages 19969--19987, Miami, Florida, USA. Association for Computational Linguistics.

\bibitem[{Xu and Zhang(2023)}]{xu2023leveraging}
Songlin Xu and Xinyu Zhang. 2023.
\newblock Leveraging generative artificial intelligence to simulate student learning behavior.
\newblock \emph{arXiv preprint arXiv:2310.19206}.

\bibitem[{Xu et~al.(2024)Xu, Zhang, and Qin}]{xu2024eduagent}
Songlin Xu, Xinyu Zhang, and Lianhui Qin. 2024.
\newblock Eduagent: Generative student agents in learning.
\newblock \emph{arXiv preprint arXiv:2404.07963}.

\bibitem[{Zeng et~al.(2024)Zeng, Yu, Gao, Meng, Goyal, and Chen}]{zeng2024llmbar}
Zhiyuan Zeng, Jiatong Yu, Tianyu Gao, Yu~Meng, Tanya Goyal, and Danqi Chen. 2024.
\newblock Evaluating large language models at evaluating instruction following.
\newblock In \emph{International Conference on Learning Representations (ICLR)}.

\bibitem[{Zhang et~al.(2019)Zhang, Kishore, Wu, Weinberger, and Artzi}]{zhang2019bertscore}
Tianyi Zhang, Varsha Kishore, Felix Wu, Kilian~Q Weinberger, and Yoav Artzi. 2019.
\newblock {BERTS}core: Evaluating text generation with {BERT}.
\newblock \emph{arXiv preprint arXiv:1904.09675}.

\bibitem[{Zhang et~al.(2024)Zhang, Zhang-Li, Yu, Gong, Zhou, Liu, Hou, and Li}]{zhang2024simulating}
Zheyuan Zhang, Daniel Zhang-Li, Jifan Yu, Linlu Gong, Jinchang Zhou, Zhiyuan Liu, Lei Hou, and Juanzi Li. 2024.
\newblock Simulating classroom education with llm-empowered agents.
\newblock \emph{arXiv preprint arXiv:2406.19226}.

\bibitem[{Zheng et~al.(2023)Zheng, Chiang, Sheng, Zhuang, Wu, Zhuang, Lin, Li, Li, Xing et~al.}]{zheng2023judging}
Lianmin Zheng, Wei-Lin Chiang, Ying Sheng, Siyuan Zhuang, Zhanghao Wu, Yonghao Zhuang, Zi~Lin, Zhuohan Li, Dacheng Li, Eric Xing, et~al. 2023.
\newblock Judging llm-as-a-judge with mt-bench and chatbot arena.
\newblock \emph{Advances in Neural Information Processing Systems}, 36:46595--46623.

\end{thebibliography}
\appendix

\section{Appendix}
\label{sec:appendix}

\subsection{Prompts for QG-SMS}

\label{appendix:prompts}

We provide the prompts used in each step of our proposed approach in Fig. \ref{fig:prompts}. For each requirement $R_d$ that we discussed in $\S$\ref{sec:dataset-construction}, we provide the following definition in the prompt:
\begin{itemize}
    \item \textbf{Item difficulty (DC)}: ``An easier question has a higher proportion of students with a correct answer.''
    \item \textbf{Item discrimination (DC)}: ``A question with higher discrimination is more effective at distinguishing between high-performing and low-performing students.''
    \item \textbf{Distractor efficiency (DE)}: ``An effective distractor is one that is chosen by at least 5\% of the students taking the quiz.''
\end{itemize}

\begin{figure*}[!t]
  \centering 

    \begin{promptboxgreen}{Step 1: Student Profile Generation}
    Given the following learning materials: \\
    \{Lecture Content / Knowledge Component Descriptions $L$\}
    
    Consider students with various understanding in a scenario where a quiz about the above learning materials is being conducted. Ensure that you generate at least 10 roles for the scenario. For each student, provide a detailed description that includes their name and their understanding of the lecture content. The distribution of understanding of lecture content must mimic that in a real classroom.
    \end{promptboxgreen}

    \begin{promptboxpink}{Step 2: Student Performance Prediction}
    Given the following learning materials: \\ 
    \{Lecture Content / Knowledge Component Descriptions $L$\} \\
    
    Below is the list of students and their reported understanding of the learning materials: \\
    \{Student Profiles from Step 1\} \\
    
    Given the following quiz questions about the lecture content: \\
    Question 1: \{Question $Q_1$\} \\ 
    Question 2: \{Question $Q_2$\} \\
    
    For each student, predict whether the student will correctly answer each question based on both the student's understanding, question's difficulty, guessing factors, etc.). If you predict “incorrect”, specify which distractor confuses the student.
    \end{promptboxpink}

    \begin{promptboxblue}{Step 3:  Evaluation}
    You are interested in finding a quiz question that satisfies the following requirement: \\
    \{Requirement $R_d$\}   \\ \\
    You are given 2 output quiz questions Output (a) and Output (b) and the analysis of the responses of each student who attempted the questions. Your task is to identify which of Output (a) and Output (b) better satisfies requirement \{$R_d$\}  based on the question content and student performance. \\
    \{Description  of $R_d$\} \\ \\
    
    \# Output (a): \{Question $Q_1$\} \\ 
    \# Output (b): \{Question $Q_2$\} \\
    
    \# Consider Students Performance: \{Predicted student performance from Step 2\} \\ 
    
    \# Which question better satisfies \{$R_d$\}, Output (a) or Output (b)? Your response should be either "Output (a)" or "Output (b)"
    \end{promptboxblue}
    \caption{Prompts for our three-step evaluation approach QG-SMS.}
    \label{fig:prompts}
\end{figure*}

\subsection{Experimental Details}
\label{app:exp-details}

\textbf{Assembling Learning Materials $L$}: We used all information about the learning materials provided in each dataset to assemble $L$. In \textit{EduAgent}, $L$ includes lecture transcripts and the textual descriptions of the slides used in the lecture. In \textit{DBE-KT}, $L$ includes the knowledge components and the associated description or definition.

\textbf{Construction of TC pairs}: For both datasets, each question is annotated with one or more related topics. From all possible pairs in the dataset, we first perform filtering. For EduAgent, each question is associated with a specific section of a lecture. We only consider pairs of questions from the same section of the same video lecture. The learning material $L$ consists of the lecture transcript and slide descriptions. For DBE-KT, each question may be linked to one or more knowledge components. We only consider pairs with exactly one differing knowledge component and at least one shared component, making the test cases more challenging. Here, the learning material L is the union of the knowledge components for the pair. Then, for each selected pair, we randomly choose one question as the label (TC = 1). The associated topic is set as the knowledge components (DBE-KT) or section (EduAgent) of this label. The other question in the pair is considered dispreferred output (TC = 0). This set up results in 286 pairs for DBE-KT and 217 pairs for EduAgent.

\textbf{Number of generated student profiles}: We are inspired by \citet{10.5555/3692070.3693667}, which simulates social scenarios for LLM alignment and selects 10 generated profiles to balance diversity (crucial for evaluating question quality) and computational efficiency. Following this approach, our prompt instructs the model to generate at least 10 roles for each learning material L. We then use all generated profiles—without filtering or augmentation—for Steps 2 and 3.

\textbf{Underlying LLM:} For all LLM-based experiments with GPT-4o, we used the \verb|gpt-4o-2024-05-13| checkpoint with the default hyperparameters. Regarding the experiments on the robustness of the student profiles (\S\ref{sec:analysis}), we ran the same prompt for Step 1 multiple times using the same default hyperparameters (temperature $= 1$). Thus, any differences in output come from the random seeds used in the API calls. 

\textbf{Baseline implementation:} For BERTScore, we use the implementation of Hugging Face \verb|evaluate|\footnote{\url{https://huggingface.co/docs/evaluate/en/index}} package (\verb|bertscore|). For KDA\footnote{\url{https://github.com/riiid/question-score}}, ChatEval \footnote{\url{https://github.com/thunlp/ChatEval}}, and QSalience\footnote{\url{https://github.com/ritikamangla/QSalience/}}, we used the code implementation provided by the authors. To obtain the expert personas for ChatEval, we utilized the AutoAgents interactive framework\footnote{\url{https://github.com/Link-AGI/AutoAgents}} given instruction $I$ as described in $\S$\ref{sec:baselines}. We used the implementation by \citealt{zeng2024llmbar}\footnote{\url{https://github.com/princeton-nlp/LLMBar}} for the remaining LLM-based evaluation approaches. 

\textbf{Comparison direction for single-scoring metrics}
\begin{itemize}
    \item Higher Topic Alignment: $\uparrow$ BERTScore, $\uparrow$ KDA, $\downarrow$ QSalience
    \item Easier question: $\downarrow$ BERTScore, $\uparrow$ KDA, $\downarrow$ QSalience
    \item Higher discrimination: $\uparrow$ BERTScore, $\downarrow$ KDA, $\uparrow$ QSalience
    \item  Higher distractor efficiency:$\uparrow$ BERTScore, $\downarrow$ KDA, $\uparrow$ QSalience
\end{itemize}

\textbf{Signficance testing}: We consider Vanilla, which simply outputs a preference given the instruction without any reasoning or intermediate steps, the base strategy for using LLMs in test item analysis. Other baselines and QG-SMS can be considered more complex strategies for using LLMs as evaluators of test items. We conduct pairwise binomial tests to examine whether each LLM-based evaluation approach (including QG-SMS) significantly improves consistent accuracy in test item analysis compared to Vanilla. We report the p-values of the binomial tests on QG-SMS improvements over the base strategy (Vanilla), as compared to all other evaluation approaches in Tbl. \ref{tab:sig-test}.

\begin{table*}[t]
\centering
\resizebox{0.63\textwidth}{!}{%
\begin{tabular}{l|ccccc}
\toprule
\textbf{Method} & \multicolumn{2}{c}{\textbf{DF}} & \multicolumn{2}{c}{\textbf{DC}} & \textbf{DE} \\
\cmidrule(lr){2-3} \cmidrule(lr){4-5} \cmidrule(lr){6-6}
\textbf{} & \textit{EduAgent} & \textit{DBE-KT} & \textit{EduAgent} & \textit{DBE-KT} & \textit{EduAgent} \\
\midrule
CoT & 0.999 & 0.996 & 0.999 & 0.999 & 0.999 \\
Metrics & 0.227 & 0.640 & 0.500 & 0.910 & 0.813 \\
Reference & 0.500 & 0.925 & 0.938 & 0.967 & 0.981 \\
Swap & 0.192 & 0.134 & 0.856 & 0.588 & 0.985 \\
ChatEval & 0.500 & 0.007 & 0.927 & 0.262 & 0.942 \\
QG-SMS & 0.007 & 0.000 & 0.252 & 0.124 & 0.093 \\
\bottomrule
\end{tabular}
}
\caption{P-values from binomial tests assessing whether the LLM-based evaluation strategy significantly improves consistent accuracy compared to the Vanilla baseline.}
\label{tab:sig-test}

\end{table*}

\subsection{Human Evaluation Details}
\label{appendix:human-eval}

\begin{table}[t]
\centering
\resizebox{0.8\linewidth}{!}{%
\begin{tabular}{lccc}
\toprule
\textbf{Method} & Diff. & Disc. & Dist. Eff.  \\

\hline
Vanilla & 73.81& \underline{56.67}& 77.08\\
CoT & 76.19& \underline{56.67}& 62.50\\
Metrics & 71.43& 53.33& \underline{81.25}\\
Reference & 73.81& 53.33& 75.00\\
Swap & 76.19& \textbf{63.33}& 77.08\\
ChatEval & 83.33& 43.33& 72.92\\
QG-SMS & \underline{85.71} & \underline{56.67}& \underline{81.25}\\
\hline
Human & \textbf{90.48}& 53.33& \textbf{87.50}\\
\bottomrule
\end{tabular}
}
\caption{Results breakdown of QG evaluation approaches and human annotators on 60 human-written question pairs. QG-SMS outperforms all baselines in terms of evaluating question difficulty and distractor efficiency, reaching closest accuracy scores to human annotators. In terms of question discrimination, QG-SMS surpasses human evaluators, reaching the second-best performance. Overall, QG-SMS shows effectiveness on three dimensions.}
\label{tab:human_eval_more}

\end{table}

\textbf{Selection of human-written question pairs:} In the \textit{EduAgent} dataset, both questions in a ($Q_1$, $Q_2$) pair comes from the same lecture. However, they can be grounded to either \textbf{the same} or \textbf{different} sections of the lecture. For example, in Tbl. \ref{tab:motivation}, $Q_1$ is relevant to the \textit{Introduction to computer vision} section, while $Q_2$ is relevant to the \textit{Computer vision history} section. To reduce the cognitive load for annotators, we opt for question pairs that are grounded to \textbf{the same section in the same lecture}. Based on this condition, we selected 60 pairs of human-written questions that exhibit differing quality: 21 pairs in the DF dimension, 15 pairs in the DC dimension, and 24 pairs in the DE dimension.

\textbf{Construction of generated question pairs}: To generate questions with varying quality regarding dimension $d$, we use the zero-shot prompts provided in Fig. \ref{fig:qg-prompts}. Using GPT-4o with the  \verb|gpt-4o-2024-05-13| checkpoint, we obtained a question bank of 360 generated questions across 5 lectures. Then, for each of the 60 human-written pairs, we construct a generated question pair grounded to the same section of the corresponding lecture and differs in the corresponding dimension $d$. 

\textbf{Instructions for annotators}: For each pair, we asked annotators to first read the section of the lecture that the pair is grounded upon before determining their preference. We provided our human annotators the same definition of each dimension $d$ in $\S$\ref{sec:statistics-definiion} and the desirable trait $R_d$ in $\S$\ref{sec:dataset-construction}. In this way, human annotators serve as another QG evaluation competitor for the human-written pairs, and provide the label for the generated-question pairs.


\begin{figure*}[!ht]
  \centering 

    \begin{promptboxgreen}{Difficulty-controlled question generation}
    Given the following learning materials: \\
    \{Lecture Content / Knowledge Component Descriptions $L$\} \\

    Generate 4-choice quiz questions to test students' understanding of the lecture. The generated questions should have diverse difficulty.
    \begin{itemize}
        \item The more difficult a question, the fewer number of students can correctly answer it.
        \item There must be 2 (two) 'easy-level' questions, 2 (two) 'medium-level' questions, and 2 (two) 'hard-level' questions.
    \end{itemize}
    \end{promptboxgreen}

    \begin{promptboxpink}{Discrimination-controlled question generation}
    Given the following learning materials: \\
    \{Lecture Content / Knowledge Component Descriptions $L$\} \\

    Generate 4-choice quiz questions to test students' understanding of the learning materials. The generated questions should have diverse discrimination ability.
    \begin{itemize}
        \item A question with high discrimination is more effective at distinguishing between high-performing and low-performing students. An example of a question with low discrimination is when neither high-performing nor low-performing students can answer the question correctly, or when all students can answer the question correctly. 
        \item There must be 2 (two) 'low-discrimination' questions, 2 (two) medium-discrimination questions, and 2 (two) 'high-discrimination' questions. 
    \end{itemize}
    \end{promptboxpink}

    \begin{promptboxblue}{Distractor-efficiency-controlled question generation}
    Given the following learning materials: \\
    \{Lecture Content / Knowledge Component Descriptions $L$\} \\

    Generate 4-choice quiz questions to test students' understanding of the lecture. The generated questions should have diverse number of effective distractors.
    \begin{itemize}
        \item An effective distractor is one that will be selected by at least 5\% of the students. 
        \item Specifically, there must be 2 (two) questions with NO effective distractors, 2 (two) questions with exactly ONE effective distractors, 2 (two) questions with exactly TWO effective distractors, and 2 (two) questions with all THREE effective distractors.
    \end{itemize}
    \end{promptboxblue}
    \caption{Prompts for generating questions with varying quality across three dimensions: difficulty, discrimination, and distractor efficiency.}
    \label{fig:qg-prompts}
\end{figure*}

\subsection{Qualitative Analysis on Limitations of Existing Evaluation Approaches}
\label{app:case-study}

We provide in Tbl. \ref{tab:case-study} the generations of all LLM-based baselines when evaluating the two questions from the case study presented in Tbl. \ref{tab:motivation}. As shown, strategies that incorporate reasoning (e.g., ChatEval, CoT, Swap) consistently prioritize questions that “apply” the concept of computer vision—indicating higher discrimination—over questions that require recalling specific statistics. Meanwhile, generation-based criteria (Metrics, Reference) are too broad and not specific enough to assess the requirement of high discrimination, likely due to being distracted by the long learning material context. Without explicitly modeling student performance, it is difficult for LLMs to "reason" about discrimination based solely on question content and materials.
\begin{table*}[!t]
\centering
\small
\begin{tabular}{@{}p{\textwidth}@{}}
\toprule
    \textbf{Learning Material} \\
    \textbf{Introduction of computer vision}: Computer vision (CV) is the field of computer science that focuses on creating digital systems \colorbox{mintgreen}{that can process, analyze}, and make sense of visual data [...]. For example, [...] \\
    \textbf{Computer vision history} [...] In 2012, a team from the \colorbox{lightpink}{University of Toronto} [...]. \\ The model, called AlexNet, [...], \colorbox{lightpink}{achieved an error rate of 16.4\%}, which overperformed all other methods at that time. [...] \\
\midrule
\textbf{Quiz Questions} \\
\textbf{$Q_1$}: Which of the following \colorbox{mintgreen}{may utilize} \colorbox{mintgreen}{computer vision techniques}? 1). Use a camera to check potential issues on the surface of products (2). Estimate the freshness of apples from pictures (3). Estimate whether a car is speeding via a camera (4). Determine whether a piece of audio is spoken by a specific person
\\ A) (1)(2)(3);   B) (1)(2)(4); C) (2)(3)(4);   D) (1)(2)(3)(4). \\ \\
\textbf{$Q_2$}: One breakthrough in computer vision happened at the \colorbox{lightpink}{University of Toronto} in 2012, which achieved an \colorbox{lightpink}{error rate of} [ ] in image classification.
\\ A) 6.4\%; B) 10.4\% C) 12.4\%  D) 16.4\%.\\
 \midrule

\textbf{Evaluation Task: Which question has higher discrimination?} \\
\textbf{\textcolor{customred}{Existing approaches: $Q_1$.}} $Q_1$ is an apply-level question, while $Q_2$ is a recall-level question. \\ 
\textbf{\textcolor{customblue}{Label based on Actual Student Performance: $Q_2$.}} Applications of CV appearing in $Q_1$ can be considered common knowledge while $Q_2$ tests a specific detail which only students who pay close attention to details may be able to answer.
\\
\midrule
\textit{CoT}: Q1 asks students to identify which scenarios \textbf{utilize computer vision techniques, which requires a deeper understanding of the lecture content and the application of computer vision}. This question effectively distinguishes between students who have a comprehensive understanding of the lecture and those who do not, as it requires them to apply their knowledge to different scenarios.

Q2, on the other hand, \textbf{asks for a specific fact from the lecture}, the error rate achieved by the University of Toronto in 2012. While this question tests recall, it does not require the same level of understanding or application of knowledge as Q1. It is more about memorizing a specific detail rather than understanding the broader concepts and applications of computer vision.
\\
\midrule
\textit{Metrics}: Does the question accurately and comprehensively reflect the specific content and structure of the lecture slides and transcripts provided in the instruction?

Does the question maintain the clarity and coherence of the information as presented in the lecture slides and transcripts? 

Does the question include all the required elements mentioned in the instruction without adding or omitting any details?
\\
\midrule
\textit{Reference}: Explain the significance of the 2012 breakthrough by the University of Toronto in the context of computer vision history, and discuss the two key factors that contributed to the success of convolutional neural networks (CNNs) at that time.
\\
\midrule
\textit{Swap}: Q1 presents a question that \textbf{requires students to apply their understanding of computer vision applications}, which were discussed in the lecture. This question effectively tests the students' ability to distinguish between tasks that can and cannot be performed using computer vision, thus having high discrimination.

Q2 \textbf{focuses on a specific fact mentioned in the lecture}, the error rate achieved by the University of Toronto in 2012. While this question tests recall of a specific detail, it does not effectively distinguish between high-performing and low-performing students as it relies more on memorization rather than understanding and application of concepts.
\\
\midrule
\textit{ChatEval}

\textbf{$Q_1$}: Relevance to Lecture Content: The question is relevant as it pertains to the applications of computer vision, which is discussed in the lecture.

Complexity and Depth of Knowledge: The question requires students to \textbf{apply their understanding of computer vision to different scenarios, which is a higher-order cognitive skill}. This makes it effective at distinguishing between high-performing and low-performing students.

Appropriateness: The question is well-designed but includes an option related to audio, which is not directly related to computer vision. This could potentially confuse students.

\\
\textbf{$Q_2$}: Relevance to Lecture Content: The question is directly relevant to the lecture content, specifically the historical breakthrough in computer vision.

Complexity and Depth of Knowledge: The question tests recall of a specific fact, which is a lower-order cognitive skill. It does not require deep understanding or application of knowledge, making it less effective at distinguishing between high-performing and low-performing students.

Appropriateness: The question is straightforward and tests a specific detail from the lecture, but it lacks complexity and does not challenge students to apply their knowledge.
\\
\bottomrule
\end{tabular}
\caption{Case Study illustrates limitations of existing evaluation metrics in test item analysis. }
\label{tab:case-study}
\end{table*}

\end{document}